\documentclass[10pt,twocolumn,letterpaper]{article}

\usepackage[final]{cvpr}              

\usepackage{graphicx}
\usepackage{amsmath}
\usepackage{amssymb}
\usepackage{booktabs}
\usepackage[ruled,vlined]{algorithm2e}
\usepackage{algorithmic}
\usepackage[dvipsnames]{xcolor}

\usepackage[pagebackref,breaklinks,colorlinks]{hyperref}

\usepackage[capitalize]{cleveref}
\crefname{section}{Sec.}{Secs.}
\Crefname{section}{Section}{Sections}
\Crefname{table}{Table}{Tables}
\crefname{table}{Tab.}{Tabs.}

\def\cvprPaperID{10245} 
\def\confName{CVPR}
\def\confYear{2022}

\newcommand\blfootnote[1]{%
  \begingroup
  \renewcommand\thefootnote{}\footnote{#1}%
  \addtocounter{footnote}{-1}%
  \endgroup
}

\begin{document}

\title{Domain-Augmented Domain Adaptation}

\author{Qiuhao Zeng \footnote[1]{}\\
Computer Science Department\\
Western University, Canada\\
{\tt\small qzeng53@uwo.ca}
\and
Tianze Luo \footnote[1]{}\\
Nanyang Technological University\\
Singapore\\
{\tt\small tianze001@ntu.edu.sg}
\and
Boyu Wang\\
Computer Science Department\\
Western University, Canada\\
{\tt\small bwang@csd.uwo.ca}
}


\maketitle

\begin{abstract}
Unsupervised domain adaptation (UDA) enables knowledge transfer from the labelled source domain to the unlabeled target domain by reducing the cross-domain discrepancy. However, most of the studies were based on direct adaptation from the source domain to the target domain and have suffered from large domain discrepancies. To overcome this challenge, in this paper, we propose the domain-augmented domain adaptation (DADA) to generate pseudo domains that have smaller discrepancies with the target domain, to enhance the knowledge transfer process by minimizing the discrepancy between the target domain and pseudo domains. Furthermore, we design a pseudo-labeling method for DADA by projecting representations from the target domain to multiple pseudo domains, and take the averaged predictions on the classification from the pseudo domains as the pseudo labels. We conduct extensive experiments with the state-of-the-art domain adaptation methods on four benchmark datasets: Office Home, Office-31, VisDA2017, and Digital datasets. The results demonstrate the superiority of our model.
\end{abstract}

\blfootnote{$*$ Both authors contributed equally to this work and are ordered alphabetically.}

\section{Introduction}
\label{sec:intro}

With the advance of deep convolutional networks, computer vision tasks such as image recognition, semantic segmentation, video processing have reached high performance with a large amount of training data. Generally, well-trained models perform well on the testing set of which the distribution bears a resemblance to that of the training set. However, directly applying such a trained model on new domains of which the data distribution is disparate from the training data, usually results in significant performance decline. Such a phenomenon is known as \textit{domain shift} \cite{wilson2020survey}, which evidently affects a direct knowledge transfer from the source domain to the target domain.

There has been a wide study on the knowledge transfer or domain adaptation from source domains to target domains \cite{pan2009survey,zhuang2020comprehensive}. The majority of such methods attempt to align the distribution of source and target domains by learning domain-invariant representations through directly minimizing the domain discrepancy of the representation distributions from the two domains \cite{shen2018wasserstein,yan2017mind}, or adversarially training the models to enforce a domain discriminator unable to distinguish features from two domains \cite{ganin2016domain,zhou2020deep}. Generally, these prevailing DA methods leverage robust and high-complexity base learners owing to their good transferable capacity brought by deep and wide architectures. 

\begin{figure}
    \centering
    \includegraphics[width=8cm]{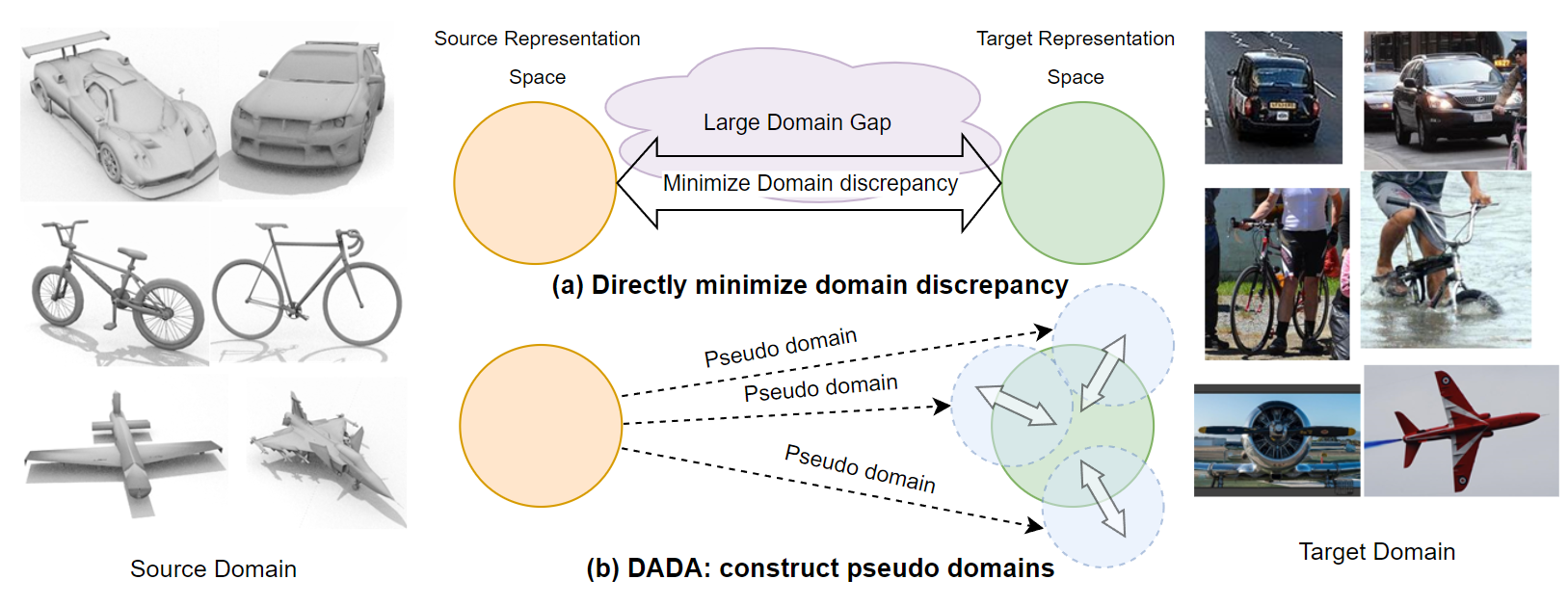}
    \caption{Overview of Domain-Augmented Domain Adaptation (DADA), compared with conventional domain discrepancy minimization method. In DADA, we generate pseudo domains that have smaller gap to the target domain, and minimize the gap between the pseudo and target domain.}
    \label{fig:intro}
\end{figure}


However, when the data from the source domain is distinct from the data in the target domain, or when the domain gap between source and target domain is relatively large, directly matching domain distributions between the two domains might cause several problems such as low convergence rate and class mismatch: a sample from source domain with label ``A" might obtain the similar representation as to the sample from the target domain with label ``B". 

In this paper, to tackle the aforementioned problem, we design an adaptive transfer learning method, named Domain-Augmented Domain Adaptation method (DADA). In DADA, as shown in Figure \ref{fig:intro}, instead of directly minimizing the domain discrepancy between the source and target domain, we expand the expressiveness of the model around the target data distribution through generating multiple pseudo domains that have small discrepancy to the target domain. Each pseudo domain is generated from the source domain by a domain generator. To maintain the distinction between each pseudo domain, each pseudo domain is associated with a domain prior, which controls the distribution of the corresponding pseudo domain. The pseudo domain can be regarded as an augmentation from the source domain in which there exists one-to-one mapping from the data in the source domain to each pseudo domain.  Therefore, each sample in every pseudo domain can be regarded as labeled training samples to train the model. However, not every pseudo domain can add value to the classifier as they may own distinct distribution from the target domain so that we need to distinguish favourable pseudo domains that can contribute to the training processes. 

\begin{figure*}[t]
    \centering
    \includegraphics[width=15cm]{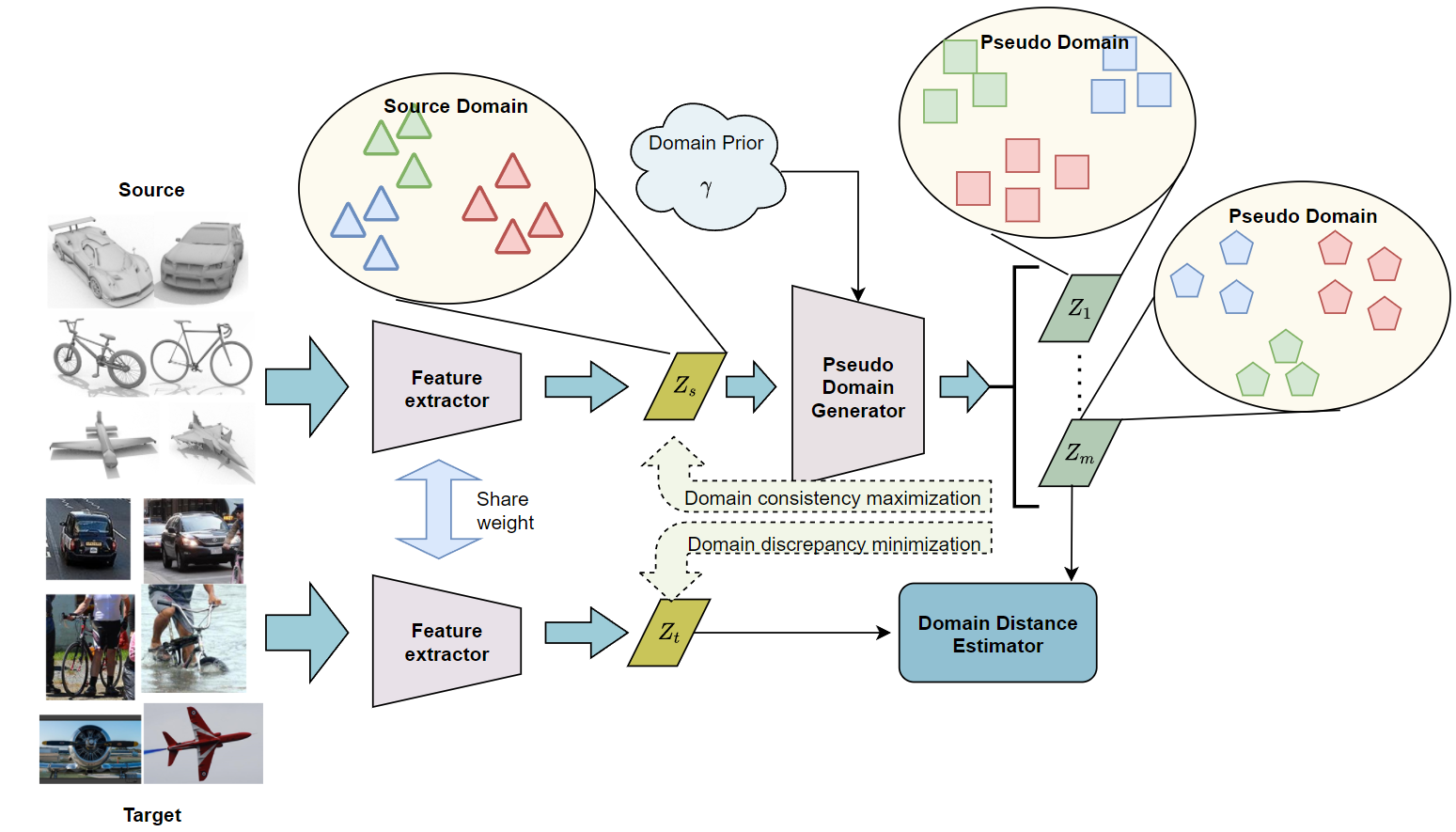}
    \caption{Illustration of the proposed domain-augmented domain adaptation (DADA) model. Firstly, the images from the source domain and the target domain are processed by the same feature extractor. Then, we sample multiple pseudo domain priors, and together with the source features $Z_s$, we apply the psuedo domain generator to generate multiple pseudo domains. In the end, the domain knowledge is transferred from the pseudo domains to the target domain. The arrows with dash lines represent the optimization objective on the correlations between the pseudo domains, the source domain and the target domain. }
    \label{fig:model}
\end{figure*}

To this end, we select favourable pseudo domains that have relatively small discrepancy to the target domain. Note that computing domain similarity generally requires a large amount of training data, which is infeasible to be done during the training process for multiple pseudo domains. Therefore, we design a domain similarity predictor to predict the similarity (e.g. Maximum Mean
Discrepancy (MMD)) between each pseudo domain and the target domain, based on the domain prior.  Based on the domain similarity, we select several pseudo domains where the pseudo domains are relatively close to the target domain, and minimize discrepancies between the pseudo domain and the target domain. 

In addition, we design a novel pseudo labeling method based on the pseudo domains. Pseudo labeling on the target domain has been proven to improve the domain adaptation capability in many related works \cite{na2021fixbi,gu2020spherical,wang2020unsupervised,li2021domain}. However, directly labeling target samples based on the classifier trained by the source domain may usually cause false labeling due to the domain shift. Instead, in DADA the representations of the target data are mapped to multiple pseudo domains where the pseudo label are taken by the average classification result of the pseudo domains. We show the superiority of this labeling method in the ablation study.

We conduct extensive experiments on four widely used datasets such as Office-Home, Office 31, VisDa and Digital datasets, to compare DADA with state-of-the-art domain adaptations methods. We also conduct extensive ablation studies and feature visualization to analyse the contribution of each component of DADA.

Our contribution can be summarized as follows:

\begin{itemize}
    \item We propose the domain-augmented domain adaptation method which can unsupervised transfer knowledge from the source domain to the target domain, through augmenting pseudo domains. 
    \item We design an appropriate pseudo labeling method for the target domain, by mapping the target data to each pseudo domain and obtaining the overall pseudo label.
    \item We conduct extensive experiments and ablation studies to test the performance of DADA compared to state-of-the-art domain adaptation methods. The experiments demonstrate the superior of DADA.
\end{itemize}

\section{Related Works}
\label{sec:rale}

\textbf{Unsupervised domain adaptation.} Unsupervised domain adaptation (UDA) deals with the problem of transferring knowledge from a labeled source domain to unlabeled target domain \cite{pan2009survey}. The mainstream of UDA methods focuses on learning domain-invariant representations through minimizing the domain discrepancy. One category of UDA method explicitly minimizes the domain discrepancy based on some common domain discrepancy measures. \cite{long2015learning,long2016unsupervised} reduce the domain discrepancy through MMD. \cite{shen2018wasserstein,lee2019sliced} minimize the Wasserstein distance between the source target domain. In DADA, we adopt MMD as the domain discrepancy measure on the source and target domain.

Another category of learning domain invariant representations is through adversarial training. This line of research has been widely explored \cite{wilson2020survey} since the influential work DANN \cite{ganin2016domain}. Adversarial learning generally trains a domain discriminator that can distinguish samples of the source domain from the target domain. Meanwhile, a feature extractor is trained to fool the discriminator to arrive at aligned features. Most works in the aforementioned categories directly minimize the discrepancy between the source and target domain, which might suffer from the problem when they encounter a large gap between the source and the target domain.


\textbf{Data augmentation.} Data augmentation can improve models' generalization capability by increasing the diversity of the training data. Conventional augmentation methods include random cropping, affine transformation (e.g. translation and rotation), horizontal flipping, etc \cite{wilson2020survey}. Recently, more augmentation techniques have been applied to domain adaptation to improve the effectiveness of knowledge transfer. MixMatch \cite{berthelot2019mixmatch} designs low-entropy labels for augmenting unlabeled instances and mixed labeled and unlabeled data for semi-supervised learning. Huang et al. \cite{huang2018auggan} design generative adversarial network (GAN) to generate augmented data for domain adaptation. DAML \cite{shu2021open} applies the Dir-Mixup to mix-up multiple source domains. FixBi \cite{na2021fixbi} includes a fixed-ratio mix-up for the source and target domain, where the mixed-up samples are the augmentation for training. TSA \cite{li2021transferable} proposes semantic augmentations to augment semantic features for the training samples. FACT \cite{xu2021fourier} utilizes Fourier transforms to mix-up the source and target domain data. ECACL \cite{li2021ecacl} designs the weak and strong data augmentation for domain adaptation.


Different from existing augmentation methods for domain adaptation, DADA proposes learnable domain augmentations, where we utilize a domain generator to learn to generate pseudo domains based on the source domain. Then we select multiple pseudo domains that have small discrepancies to the target domain as our augmentations to enhance the knowledge transfer to the target domain.


\section{Methodology}
\label{sec:method}

For UDA classification, we denote the source domain as $\mathcal{D}_s = \{(x_i^s, y_i^s)\}_{i=1}^{N_s}$ with $N_s$ labeled samples, where $x_i^s$ and $y_i^s$ denote the $i^{\text{th}}$ sample and its label respectively. We denote the target domain as $\mathcal{D}_t = \{(x_j^t)\}_{j=1}^{N_t}$ with $N_t$ samples. Both domains share the same label space $Y=\{1,2,...,K\}$ with in total $K$ classes. The samples of the source domain and target domain are drawn from different distributions $\mathcal{P}_s$ and $\mathcal{P}_t$, such that $\mathcal{P}_s \neq \mathcal{P}_t$ and our goal is to train the model on the source domain and generalize to the target domain.


\subsection{Model Overview}

The overall model architecture is shown in Figure \ref{fig:model}. Firstly, we input the samples from both the source and target domain to the model, where the data are processed by a feature extractor (e.g. ResNet-50, ResNet-101, etc).\footnote{We denote the feature extractor as the backbone model in some following context.} Subsequently, we obtain the representations from the source domain and the target domain with the distribution $\mathcal{P}_{s}$ and $\mathcal{P}_{t}$ respectively. In the next step, we apply a pseudo domain generator to generate multiple pseudo domains, where each domain associates with a unique domain prior. The domain prior is sampled from a uniform distribution. Therefore, we can form plenty of pseudo domains, where there exists some pseudo domains that have a low discrepancy between the target domain, such that it is also easier for us to transfer knowledge from such pseudo domains to the target domain. To select appropriate pseudo domains, we use a domain discrepancy estimator, which ,based on the domain prior, estimates the pseudo and target domains' discrepancy. To transfer knowledge to the target domain, we design a set of contrastive loss functions to ensure the knowledge can be transferred to the target domain. 


\subsection{Pseudo Domain Generation}

The key component of DADA is the domain generation network $f_{g}$ that generates multiple pseudo domains based on the source domain. For each pseudo domain generation, first, we randomly sample a set of $m$ domain priors $\{\gamma_j\}_{j=1}^{m}$ from a discrete uniform distribution $\mathcal{U}$ that contains evenly spaced numbers over $[-0.5, 0.5]$ with the constant interval 0.01, where $\gamma_j \in \mathbb{R}^1$ denotes the $j^{\text{th}}$ sampled domain prior. $m$ is a tunable hyper-parameter that controls the number of pseudo domains where we generate in each batch. Each domain prior $\gamma_{j}$ is associated with a specific pseudo domain $\mathcal{D}_{\gamma_j}$. The domain generator can be denoted as a mapping function: $f_{g} : \mathbb{R}^d \times \mathbb{R}^1 \rightarrow \mathbb{R}^d$. In practice, we concatenate each domain prior $\gamma_j$ with the features that output from the feature extractor from the source domain, and we denote the new representation as $H_j = {h_{1}^j,...,h_{B}^j}\in \mathbb{R}^{|B|*(d+1)}$, where $|B|$ denotes the batch size and $h_{i}^j\in{\mathbb{R}^{d+1}}$ is the feature of the $i^{\text{th}}$ sample with the prior $\gamma_j$.



We apply a multilayer perceptron network to parameterize the domain generator and the output of the domain generator is a batch of samples of the pseudo domain that is associated with the domain prior $\gamma_j$. We denote the output as $Z_j = \{z_{1}^j,...,z_{B}^j\}\in \mathbb{R}^{|B|*d}$. With $m$ priors, we can obtain $m$ distinct batches $\{Z_j\}_{j=1}^m$ of representations from the pseudo domains $\{\mathcal{D}_{\gamma_1},...,\mathcal{D}_{\gamma_m}\}$. 

\subsubsection{Domain consistency loss}


To ensure each sample from the generated pseudo domain maintains the same information as the original sample from the source domain, we design a domain consistency loss, to regularize the representation of the pseudo domains. To this end, since each sample $x^s_i$ in the source domain with representation $z^s_i$ acquires a unique corresponding representation $\hat{z}_i^j$ from the pseudo domain $\mathcal{D}_{\gamma_{j}}$, we apply the contrastive learning to build strong relationships between the paired sampled $z^s_i$ and $\hat{z}_i^j$: maximize the correlation for the positive pair $(z^s_i, \hat{z}_i^j)$ and minimize the correlation for negative pairs $(z^s_k, \hat{z}_i^j)$, where the negative samples $z^s_k$ are the data from the same batch that have different labels as the current data. Therefore, the contrastive learning can enhance inter-domain consistencies for the pseudo domains and the source domain, such that, same as the source domain, the representations of the data in the pseudo domain with the same labels can be grouped together. Consequently, for downstream tasks, a classifier can easily distinguish the data from different classes.




The domain consistency loss for the pseudo domain $\mathcal{D}_{\gamma_j}$ can be written as 

\begin{equation}
    \ell_c^j = -\sum_{i=1}^{N} \frac{\text{sim}(z_i, \hat{z}_i^j)}{\text{sim}(z_i, \hat{z}_i^j) + \sum_{k=1, y_k\neq y_i}^{|B|} \text{sim}(z_k, \hat{z}_i^j)},
    \label{eq:consistency}
\end{equation}

\noindent where $z_i$, $z_k$ and $\hat{z}_{i}^j$ are samples from the batch $B$ with batchsize of $N$, $\text{sim}$ denotes the similarity function, where we apply a bilinear function to compute the similarity of two representations. It is worth noting that a cosine or inner product similarity measure may not work in our setting, as those measures highly restrict the distribution of the pseudo domains, so that we may hardly find pseudo domains that have small discrepancy with the target domain when the discrepancy between the source and target domain is large. We take the $z_i$ and $\hat{z}_{i}^j$ which are from the same data sample $x_i$ as the positive pair, while considering representations with different labels as negative pairs. 

Through the loss function in (\ref{eq:consistency}), the representations from pseudo domains can be regularized by the corresponding representations from the source domains, where the mutual information for the corresponding positive pairs can be maximized by (\ref{eq:consistency}). 






\subsection{Domain Discrepancy Estimation and Minimization}

\subsubsection{Domain discrepancy estimation} Pseudo domains $\{\mathcal{D}_{\gamma_j}\}_{j=1}^m$ are generated by the domain generator based on the domain prior, where some pseudo domains are generated with similar distributions as the target domain, while others are dissimilar to the target domain. We hope to select the pseudo domains that have small discrepancy (i.e. similar) to the target domain, since transferring knowledge from such pseudo domains to the target domain is easier comparing to random pseudo domains. However, estimating the discrepancy between two domains requires a large amount of training data, as computing the MMD based on a single batch of data does not accurately reflect the MMD of two domains. Thus, it is unfavorable for conventional domain adaptation models to concurrently estimate domain discrepancies and perform domain adaptation, since domain adaptation procedure changes the result of domain discrepancies.


Therefore, we design an explicit domain discrepancy estimator $\delta$. Since each pseudo domain is generated based on a specific domain prior, we can directly estimate the discrepancy (e.g. MMD) through regression based on the domain prior as the following loss function:


\begin{equation}
    \ell_{dde}(\gamma,Z_t) = (\text{MMD}(Z_\gamma,Z_t) - \delta(\gamma))^2,
    \label{eq:estimator}
\end{equation}

\noindent where $Z_\gamma$ denotes the representations from the pseudo domain $\mathcal{D}_{\gamma}$ with prior $\gamma$, and $\delta$ is the discrepancy estimator, where we parameterize it with a multilayer perceptron network. By minimizing the error in (\ref{eq:estimator}), the discrepancy estimator can gradually estimate the MMD between any generated pseudo domains and the target domain. 

\subsubsection{Inter-pseudo-domain discrepancy maximization} 

For the $m$ pseudo domains, we hope each pseudo domain can form a distinct distribution in the latent space. The motivation behind this is that if each pseudo domain forms a distribution that is near to the target domain and distinct from each other, we can have a generalized representation for the space that surrounds the target domain. In contrast, if all the pseudo domains share a similar distribution on the feature representation, it will be meaningless to generate $m$ pseudo domains instead of a single pseudo domain. 


To this end, we try to maximize the discrepancy between pseudo domains by adding a regularization term to the MMD between pseudo domains as follows:

\begin{equation}
    \ell_{ddm} = \sum_{i,j\in m, i\neq j}-\text{MMD}(Z_{\gamma_i}, Z_{\gamma_{j}}),
    \label{eq:seperation}
\end{equation}

\noindent where this regularization term serves as a distribution separator to spread the pseudo domain representations to cover the spaces that are close to the target domain.

\subsubsection{Domain discrepancy minimization}


The next step is to transfer knowledge from the appropriate pseudo domains (selected by the domain estimator) to the target domain by minimizing the discrepancy between them. To this end, firstly, we assign a weight $w_j$ ($j\in \{1,...,m\}$) to each pseudo domain $\mathcal{D}_{\gamma_j}$, where the $w_j$ reflects the discrepancy between $\mathcal{D}_{\gamma_j}$ and $\mathcal{D}_t$, and is computed according to the softmax formula:

\begin{equation}
    w_{j} = \frac{\exp{\delta(\gamma_j)}}{\sum_{j=1}^{m}\exp{\delta(\gamma_j)}}.
\end{equation}

Therefore, in each batch, given $Z_s$, $Z_t$, $\{\gamma_j\}_{j=1}^{m}$ and representations $\{Z_j\}_{j=1}^{m}$ from the pseudo domains, the loss function to transfer knowledge from the pseudo domains to the target domain can be written as 

\begin{equation}
    \ell_{\mathcal{D}} = \sum_{j=1}^{m}w_j\text{MMD}(Z_t,Z_j).
\end{equation}



\subsection{Pseudo Labeling and (Self-)Supervised Loss}

Psuedo labeling has been shown to improve the model performance in recent DA works \cite{na2021fixbi,gu2020spherical,wang2020unsupervised,li2021domain}. Most of the existing pseudo labeling methods train the classifier by the source data and directly apply it to the training samples from the target domain to obtain pseudo labels. However, such pseudo labeling highly depends on the discrepancy between the source and target domain, and might be inaccurate when the domain gap is relatively large. 

Therefore, in our model, we design the reverse pseudo labeling method, by projecting the data from the target domain to multiple pseudo domains, and labeling the target domain samples through the classifier trained from the pseudo domain. We denote the projection function as $\xi:Z_{t}\times \gamma_j \rightarrow \hat{Z}_{\gamma_{j}}$. In general, for a well trained model, the discrepancy between the target and pseudo domain is relatively small, and applying non-linear mapping may disturb the distribution after mapping and may even increase the discrepancy. Therefore, we compute the domain centre, i.e. mean of the representations from both domains, and directly add the difference of the mean to the data in the target domain. $\hat{Z}_{\gamma_{j}}$ can be expressed as:

\begin{equation}
    \hat{Z}_{\gamma_{j}} = Z_t + (\frac{1}{n}\sum_{i=1}^n z_i^{\gamma_{j}} - \frac{1}{n}\sum_{i=1}^n z_i^{t}).
    \label{eq:reverse}
\end{equation}



Then we compute the probability of the label $p_j = p(\hat{y}|\hat{Z}_{\gamma_{j}})$for $\hat{Z}_{\gamma_{j}}$ based on the classifier. Therefore, for $m$ pseudo domains, we can obtain $m$ set of probabilities. Then, we compute the overall probability of the labels $\hat{p} = \frac{1}{m}\sum_{j=1}^m p_j$. Base on the probability, we assign the label with the highest probability as the pseudo label for the samples from the target domain. We denote the pseudo label as $y_t$.

Therefore, the self-supervised loss for the target domain can be written as:

\begin{equation}
    \ell_{sup}^t = -\frac{1}{N}\sum_{i=1}^{N}\hat{y}_i^t \text{log}(q(y|x_{i}^t))+\beta \ell_{MI},
\end{equation}

\noindent where $\hat{y}_i^t$ is the predicted pseudo label. $q(y|x_{i}^t)\in \mathbb{R}^C$ denotes the predicted probability on each class, and $C$ is the number of classes. $\ell_{MI}$ is a regularization term adopted from \cite{li2021transferable} which regularizes the distribution of the target prediction by encouraging predictions on high confident classes (i.e. classes with high predicted probability) and enhancing balanced prediction on each classes. Typically, $\ell_{MI}=\hat{q}(y|x^t)\text{log }\hat{q}(y|x^t) - \frac{1}{N}q(y|x_i^t)\text{log }q(y|x_i^t)$, where $\hat{q}(y|x^t) = \frac{1}{N}\sum_{i=1^N}q(y|x_i^t)$ is the average prediction of the batch. $\beta$ is a tunable parameter where we tested $\beta$ value from $\{0.1,0.2,..1.0\}$. Meanwhile, to train a robust classifier, we need the supervised loss for the source domain and the pseudo domains as follows:

\begin{equation}
    \ell_{sup}^s = -\frac{1}{N}\sum_{i=1}^{N}y_i^s \left (\text{log}(q(y|x_{i}^s))+\sum_{j=1}^m \text{log}(q(y|x_{i}^{\gamma_j})) \right ).
\end{equation}




\subsection{Loss functions and Overall Algorithm}

The overall loss functions consist of the supervised loss for the source and pseudo domains, pseudo domain consistency loss (\ref{eq:consistency}), domain discrepancy losses, as well as target domain self-supervised loss. Thus the overall loss function can be derived as:




\begin{equation}
    \mathcal{L} = \ell_{sup}^s + \lambda_1 \ell_{sup}^t + \lambda_2 \ell_c + \lambda_3 \ell_{\mathcal{D}}+\lambda_4 \ell_{ddm} + \lambda_5 \ell_{dde},
\end{equation}

\noindent where $\lambda_1, \lambda_2, \lambda_3, \lambda_4, \lambda_5$ are hyper-parameters to adjust the weight of each loss components. Algorithm \ref{alg:1} summarizes the proposed method.

\begin{algorithm}[!t]
\caption{Main Algorithm.}
\label{alg:1}
\begin{algorithmic}[1]
\STATE {\bfseries Input:} batch size $N$, number of pseudo domains $m$, constant $lr$
\FOR{sampled minibatch $\{(x_{i}^s, y_{i}^s)\}_{i=1}^N$, $\{x_{i}^t\}_{i=1}^N$ \textbf{do}} 
    \STATE get $Z_s$ and $Z_t$ from the feature extractor
    \STATE sample $m$ pseudo domain priors $\{\gamma\}_{j=1}^m\sim \mathcal{U}\{-0.5,0.5\}$
    \STATE generate pseudo domains $\{\mathcal{D}_{\gamma_j}\}_{j=1}^m$ by the domain generator
    \STATE compute the domain consistancy loss $\ell_c$
    \STATE compute and estimate domain discrepancies $\text{MMD}(Z_{\gamma_{j}},Z_t)$ and $\ell_{dde} (\gamma,Z_t)$
    \STATE compute inter-pseudo discrepancy $\ell_{ddm}$ for pseudo domains
    \STATE compute domain discrepancy loss $\ell_{\mathcal{D}}$
    \STATE compute the pseudo label $\hat{Z}$
    \STATE compute the supervised loss $\ell_{sup}^s$ and $\ell_{sup}^t$ 
    \STATE $\mathcal{L} = \ell_{sup}^s + \lambda_1 \ell_{sup}^t + \lambda_2  \ell_{\mathcal{D}}+\lambda_3 \ell_{c} + \lambda_4 \ell_{ddm} + \lambda_5 \ell_{dde}  $
    \STATE update networks $\theta$ to minimize $\mathcal{L}$
    
        
    
\ENDFOR
\STATE Return networks $\theta$
%
\end{algorithmic}
\end{algorithm}

\section{Experiment}

\begin{table*}[htbp]
  \small
  \setlength{\abovecaptionskip}{0.cm}
  \setlength{\belowcaptionskip}{0.cm}
  \centering
  \caption{Accuracy (\%) on Office-Home for UDA (ResNet-50).}
  \setlength{\tabcolsep}{0.4mm}{
    \begin{tabular}{|l|cccccccccccc|c|}
    \hline
    Method & Ar$\rightarrow$Cl & Ar$\rightarrow$Pr & Ar$\rightarrow$Rw & Cl$\rightarrow$Ar & Cl$\rightarrow$Pr & Cl$\rightarrow$Rw & Pr$\rightarrow$Ar & Pr$\rightarrow$Cl & Pr$\rightarrow$Rw & Rw$\rightarrow$Ar & Rw$\rightarrow$Cl & Rw$\rightarrow$Pr & Avg \\
    \hline
    \hline
    JAN \cite{JAN} & 45.9 & 61.2 & 68.9 & 50.4 & 59.7 & 61.0 & 45.8 & 43.4 & 70.3 & 63.9 & 52.4 & 76.8 & 58.3 \\
    TAT \cite{TAT} & 51.6 & 69.5 & 75.4 & 59.4 & 69.5 & 68.6 & 59.5 & 50.5 & 76.8 & 70.9 & 56.6 & 81.6 & 65.8 \\
    TPN \cite{TPN} & 51.2 & 71.2 & 76.0 & 65.1 & 72.9 & 72.8 & 55.4 & 48.9 & 76.5 & 70.9 & 53.4 & 80.4 & 66.2 \\
    ETD \cite{ETD} & 51.3 & 71.9 & \textbf{85.7} & 57.6 & 69.2 & 73.7 & 57.8 & 51.2 & 79.3 & 70.2 & 57.5 & 82.1 & 67.3 \\
    SymNets \cite{SymNets} & 47.7 & 72.9 & 78.5 & 64.2 & 71.3 & 74.2 & 64.2 & 48.8 & 79.5 & 74.5 & 52.6 & 82.7 & 67.6 \\
    BNM \cite{BNM} & 52.3 & 73.9 & 80.0 & 63.3 & 72.9 & 74.9 & 61.7 & 49.5 & 79.7 & 70.5 & 53.6 & 82.2 & 67.9 \\
    MDD \cite{MDD} & 54.9 & 73.7 & 77.8 & 60.0 & 71.4 & 71.8 & 61.2 & 53.6 & 78.1 & 72.5 & 60.2 & 82.3 & 68.1 \\
    GSP \cite{GSP} & 56.8 & 75.5 & 78.9 & 61.3 & 69.4 & 74.9 & 61.3 & 52.6 & 79.9 & 73.3 & 54.2 & 83.2 & 68.4 \\
    GVB-GD \cite{GVB} & 57.0 & 74.7 & 79.8 & 64.6 & 74.1 & 74.6 & 65.2 & 55.1 & 81.0 & 74.6 & 59.7 & 84.3 & 70.4 \\
    TSA\cite{li2021transferable} & 57.6 & 75.8 & 80.7 & 64.3 & 76.3 & 75.1 & \textbf{66.7} & \textbf{55.7} & 81.2 & \textbf{75.7} & \textbf{61.9} & 83.8 & 71.2 \\
    \hline
    Ours & \textbf{58.9} &  \textbf{79.5} & 82.2 & \textbf{66.3} & \textbf{78.2} & \textbf{ 78.2} & 65.9 & 53 & \textbf{81.6} & 74.5 & 60.2 & \textbf{85.1} & \textbf{72.0}\\
    \hline
    \end{tabular}}
  \label{tab:Office-Home}
\end{table*}

\begin{table*}[h]
    \centering
\begin{tabular}{|c|cccccccccccc|c|}
\hline Method & plane & bcycl & bus & car & horse & knife & mcycl & person & plant & sktbrd & train & truck & Avg. \\
\hline ResNet [12] & $55.1$ & $53.3$ & $61.9$ & $59.1$ & $80.6$ & $17.9$ & $79.7$ & $31.2$ & $81.0$ & $26.5$ & $73.5$ & $8.5$ & $52.4$ \\
DAN [24] & $87.1$ & $63.0$ & $76.5$ & $42.0$ & $90.3$ & $42.9$ & $85.9$ & $53.1$ & $49.7$ & $36.3$ & $85.8$ & $20.7$ & $61.6$ \\
DANN [9] & $81.9$ & $77.7$ & $82.8$ & $44.3$ & $81.2$ & $29.5$ & $65.1$ & $28.6$ & $51.9$ & $54.6$ & $82.8$ & $7.8$ & $57.4$ \\
MinEnt [11] & $80.3$ & $75.5$ & $75.8$ & $48.3$ & $77.9$ & $27.3$ & $69.7$ & $40.2$ & $46.5$ & $46.6$ & $79.3$ & $16.0$ & $57.0$ \\
MCD [36] & $87.0$ & $60.9$ & $83.7$ & $64.0$ & $88.9$ & $79.6$ & $84.7$ & $76.9$ & $88.6$ & $40.3$ & $83.0$ & $25.8$ & $71.9$ \\
ADR [35] & $87.8$ & $79.5$ & $83.7$ & $65.3$ & $92.3$ & $61.8$ & $88.9$ & $73.2$ & $87.8$ & $60.0$ & $85.5$ & $32.3$ & $74.8$ \\
SWD [14] & $90.8$ & $82.5$ & $81.7$ & $70.5$ & $91.7$ & $69.5$ & $86.3$ & $77.5$ & $87.4$ & $63.6$ & $85.6$ & $29.2$ & $76.4$ \\
CDAN+E [25] & $85.2$ & $66.9$ & $83.0$ & $50.8$ & $84.2$ & $74.9$ & $88.1$ & $74.5$ & $83.4$ & $76.0$ & $81.9$ & $38.0$ & $73.9$ \\
AFN [40] & $9 3 . 6$ & $61.3$ & $\mathbf{8 4 . 1}$ & $\mathbf{7 0 . 6}$ & $\mathbf{9 4 . 1}$ & $79.0$ & $\mathbf{9 1 . 8}$ & $79.6$ & $89.9$ & $55.6$ & $89.0$ & $24.4$ & $76.1$ \\
BNM [7] & $89.6$ & $61.5$ & $76.9$ & $55.0$ & $89.3$ & $69.1$ & $81.3$ & $65.5$ & $90.0$ & $47.3$ & $\mathbf{8 9 . 1}$ & $30.1$ & $70.4$ \\
MSTN+DSBN [3] & $94.7$ & $86.7$ & $76.0$ & $72.0$ & $95.2$ & $75.1$ & $87.9$ & $\textbf{81.3}$ & $\textbf{91.1}$ & $68.9$ & $88.3$ & $45.5$ & $80.2$ \\
\hline Ours & \textbf{95.3} & \textbf{87.7} & 80.9 & 56.8 & 93.1 & \textbf{84.1} & 81.4 & 78.3 & 90.3 & \textbf{82.3} & 87.2 & \textbf{52.8} & \textbf{80.9}\\
\hline
\end{tabular}
    \caption{Accuracy(\%) on VisDA-2017 dataset for unsupervised domain adaptation (ResNet-101).}
    \label{tab:visda}
\vspace{-0.2cm}
\end{table*}

\begin{table}[htbp]
  \small
    \setlength{\abovecaptionskip}{0cm}
  \setlength{\belowcaptionskip}{0cm}
  \centering
  \caption{Accuracy (\%) on Office-31 for UDA (ResNet-50).}
    \setlength{\tabcolsep}{0.5mm}{
    \begin{tabular}{|l|cccccc|c|}
    \hline
    Method & A$\rightarrow$W & D$\rightarrow$W & W$\rightarrow$D & A$\rightarrow$D & D$\rightarrow$A & W$\rightarrow$A & Avg \\
    \hline
    \hline
    ADDA \cite{ADDA} & 86.2 & 96.2 & 98.4 & 77.8 & 69.5 & 68.9 & 82.9 \\
    JAN \cite{JAN} & 85.4 & 97.4 & 99.8 & 84.7 & 68.6 & 70.0 & 84.3 \\
    ETD \cite{ETD} & 92.1 & \textbf{100.0} & \textbf{100.0} & 88.0 & 71.0 & 67.8 & 86.2 \\
    MCD \cite{MCD} & 88.6 & 98.5 & \textbf{100.0} & 92.2 & 69.5 & 69.7 & 86.5 \\
    BNM \cite{BNM} & 91.5 & 98.5 & \textbf{100.0} & 90.3 & 70.9 & 71.6 & 87.1 \\
    DMRL \cite{DMRL} & 90.8 & 99.0 & \textbf{100.0} & 93.4 & 73.0 & 71.2 & 87.9 \\
    SymNets \cite{SymNets} & 90.8 & 98.8 & \textbf{100.0} & 93.9 & 74.6 & 72.5 & 88.4 \\
    TAT \cite{TAT} & 92.5 & 99.3 & \textbf{100.0} & 93.2 & 73.1 & 72.1 & 88.4 \\
    MDD \cite{MDD} & 94.5 & 98.4 & \textbf{100.0} & 93.5 & 74.6 & 72.2 & 88.9 \\
    GVB-GD \cite{GVB} & \textbf{94.8} & 98.7 & \textbf{100.0} & 95.0 & 73.4 & 73.7 & 89.3 \\
    GSP \cite{GSP} & 92.9 & 98.7 & 99.8 & \textbf{94.5} & 75.9& 74.9 & 89.5 \\
    \hline
    Ours & 94.5& 98.7& \textbf{100.0} &92.6 & \textbf{76.5}& \textbf{75.0} & \textbf{89.6}\\
    \hline
    \end{tabular}}
  \label{tab:Office-31}
\end{table}

\begin{table}[htbp]\scriptsize
  \small
    \setlength{\abovecaptionskip}{0cm}
  \setlength{\belowcaptionskip}{0cm}
  \centering
  \caption{Accuracy (\%) on Digital Datasets for UDA.}
  \setlength{\tabcolsep}{0.9mm}{
    \begin{tabular}{|l|ccc|c|}
    \hline
    Method & M $\rightarrow$ U &  U $\rightarrow$ M &  SV $\rightarrow$ M  & Avg \\
    \hline
    \hline
    ADDA \cite{ADDA} & 89.4$\pm$0.2 & 90.1$\pm$0.8 & 76.0$\pm$1.8 & 96.3$\pm$0.4  \\
    PixelDA \cite{DADA} & 95.9$\pm$0.7 & - & - & -  \\
    DIFA \cite{AFAUDA} & 92.3$\pm$0.1 & 89.7$\pm$0.5 & 89.7$\pm$2.0 & $90.6 \pm 0.9$ \\
    UNIT \cite{UNIT} & 95.9$\pm$0.3 & 93.6$\pm$0.2 & 90.5$\pm$0.3 & $93.3 \pm 0.3$ \\
    CyCADA \cite{CyCADA} & 95.6$\pm$0.2 & 96.5$\pm$0.1 & 90.4$\pm$0.4 & $94.2 \pm 0.2$ \\
    TPN \cite{TPN} & 92.1$\pm$0.2 & 94.1$\pm$0.1 & 93.0$\pm$0.3 & $93.1 \pm 0.2$ \\
    DM-ADA \cite{DM-ADA} & \textbf{96.7}$\pm$\textbf{0.5} & 94.2$\pm$0.9 & 95.5$\pm$1.1 & $95.5 \pm 0.8$ \\
    MCD \cite{MCD} & 96.5$\pm$0.3 & 94.1$\pm$0.3 & 96.2$\pm$0.4 & $95.6 \pm 0.3$  \\
    ETD \cite{ETD} & 96.4$\pm$0.3 & 96.3$\pm$0.1 & \textbf{97.9}$\pm$\textbf{0.4} & $96.9 \pm 0.3$ \\
    DMRL \cite{DMRL} & 96.1$\pm$0.2 & \textbf{99.0$\pm$0.1} & 96.2$\pm$0.4 & \textbf{97.1}$\pm $\textbf{0.3} \\
    \hline
    Ours & 96.6 $\pm$0.3& 97.5$\pm$0.1 & 97.2$\pm$0.3 & \textbf{97.1}$ \pm $\textbf{0.2} \\
    \hline
    \end{tabular}}
  \label{tab:digital}
  \vspace{-3mm}
\end{table}

\subsection{Datasets}

\textbf{Office-31} \cite{saenko2010adapting} is a classical cross-domain benchmark, including images in 31 classes drawn from 3 distinct domains: Amazon (A), Webcam (W) and DSLR (D).

\textbf{Office-Home} \cite{venkateswara2017deep} is a challenging benchmark for domain adaptation, which contains 15,500 images in 65 classes drawn from 4 distinct domains: Artistic (Ar), Clip Art (Cl), Product (Pr), and Real-World (Rw).

\textbf{VisDA-2017} \cite{peng2017visda} is a large-scale dataset for dataset for 2017 Visual Domain
Adaptation Challenge \footnote{http://ai.bu.edu/VisDA-2017/}. It contains over 280K images across 12 categories. Following \cite{saito2018maximum}, we choose the synthetic domain with 152,397 images as source and the realistic domain with 55,388 images as target.

\textbf{Digital Datasets} contain 4 standard digital datasets: MNIST \cite{MNIST}, USPS \cite{USPS} and Street View House Numbers (SVHN) \cite{svhn}. All of these datasets provide number images from 0 to 9. We construct four transfer tasks: MNIST to USPS (M $\rightarrow$ U), USPS to MNIST (U $\rightarrow$ M), SVHN to MNIST (SV $\rightarrow$ M) and SYN to MNIST (SY $\rightarrow$ M).



\begin{table*}[t]
\begin{center}
\caption{Ablation results (\%) of investigating the effects of our components on Office-31.}
\label{tab:ablation}
\begin{tabular}{|c|c|c|c|c||c|c|c|c|c|c|c|}
\hline
$\ell_{\text{MMD only}}$ & $\ell_{sup}^t$ & $\ell_{ddm}$ & no mapping & mapping & A$\rightarrow$W & D$\rightarrow$W & W$\rightarrow$D & A$\rightarrow$D & D$\rightarrow$A &W$\rightarrow$A & Avg\\
\hline \hline
\checkmark& \checkmark &	 &	 &	& 82.0&	96.9&	99.1&	79.7&	68.2&	67.4& 82.2\\
&	\checkmark&	 && 	&86.5&	98.4&	\textbf{100.0}&	85.5&	71.4&	71.5& 85.5\\
& \checkmark&	\checkmark&	 &	& 90.1&	98.5&	\textbf{100.0}&	88.4&	72.5&	72.5& 87.0\\
& \checkmark&	\checkmark& \checkmark&	& 92.3&	98.6&	\textbf{100.0}&	90.4&	76.3& 74.1&	88.6\\
& \checkmark&	\checkmark& &	\checkmark& \textbf{94.5}&	\textbf{98.7}&	\textbf{100.0}&	\textbf{92.6}&	\textbf{76.5}& \textbf{75.0}&	\textbf{89.6}\\
\hline
\end{tabular}
\end{center}
\vspace{-5mm}
\end{table*}

\begin{table}[htbp]
  \small
  \vspace{-0.2cm}
    \setlength{\abovecaptionskip}{0cm}
  \setlength{\belowcaptionskip}{0cm}
  \centering
  \caption{Accuracy (\%) on Office-31 for UDA (ResNet-50) with different number of pseudo domains: $m \in \{1,...,7\}$.}
    \setlength{\tabcolsep}{1mm}{
    \begin{tabular}{|c||cccccc|c|}
    \hline
    $m$ value & A$\rightarrow$W & D$\rightarrow$W & W$\rightarrow$D & A$\rightarrow$D & D$\rightarrow$A & W$\rightarrow$A & AVG\\
    \hline
    1 & 93.5 &98.2& 99.2& 93.1& 75.2& 74.2  & 88.8 \\
    2 & 92.8 &98.1& 99.3& 93.0& 75.5& 75.3  & 88.9\\
    3 & 93.6 & 98.6 & 99.8 & 92.6 & 75.8 & 75.2  & 89.3 \\
    4 & 94.2& 99.0& 100&  94.2&  76.3& 74.8 & \textbf{89.8}\\
    5 & 92.7 & 98.4 & 99.6 & 92.4 & 75.3 & 75.0 & 88.9 \\
    6 &  93.8 & 98.8& 99.7& 93.8 & 76.2 & 75.2 & 89.6\\
    7 & 94.1 & 98.5 & 99.8 & 93.4 & 76.0 & 75.4 & 89.5 \\
    \hline
    \end{tabular}}
  \label{tab:pseudo_m}
  \vspace{-3mm}
\end{table}

\subsection{Implementation details}

We follow the standard UDA protocol by training our model on all the labeled source data and unlabeled target data. For Office-Home and Office-31, we use ResNet-50 as our backbone, and we use ResNet 101 to train VisDA-2017 dataset. For the Digital datasets, we employ the same network as \cite{li2021transferable,li2019joint} and we train the network from scratch. We use a stochastic gradient descent optimizer with the momentum of 0.9, learning rate of 0.001, and weight decay of $5\times 10^{-4}$. For the hyper-parameters of the loss functions, we set $\lambda_1 = 0.5, \lambda_2 = 0.1, \lambda_3 = 0.01, \lambda_4 = 0.01, \lambda_5=0.1$, number of pseudo domains $m=4$. We set the batchsize of 64 for VisDA-2017, Office Home and Office-31, and batchsize of 32 for the digital datasets. We train our model for 80 epochs with 10 warm-up epochs. The entire model is implemented using PyTorch \footnote{The source code is contained in the Supplementary. We will release the source code to the community upon the acceptance of this paper.}. To evaluate the performance of DADA, we use various source-only models and state-of-the-art UDA models as our baselines to validate the effectiveness of DADA, such as GSP \cite{GSP} GVB-GD \cite{GVB} TSA \cite{li2021transferable}, ect.



\subsection{Comparison with the State-of-the-Art models}

\textbf{Office-Home} The results of the comparative performance on the dataset Office-Home with ResNet-50 are shown in Table \ref{tab:Office-Home}. DADA demonstrates particularly strong performance in the domain adaptation task between real-world domain and artificial domain, such as Pr $\rightarrow$ Rw, Cl $\rightarrow$ Rw, Rw $\rightarrow$ Pr. Especially when the real-world domain serves as the target domain, e.g. Cl $\rightarrow$ Rw and Pr $\rightarrow$ Rw, DADA achieves highest or second highest performance. Since the data distribution form the real-world usually has a higher variance than the variance of the artificial data, applying multiple pseudo domains can generalize the classification capability to the pseudo space that in general covers the target representation space. Such a phenomenon makes DADA suitable to be applied in real-world image classification for adapting knowledge from any synthetic/artificial data to related unlabeled real-world data. Overall, DADA has achieved an average performance of 72.0 which outperforms the baselines. Especially, compared to SOTA augmentation methods, such as TSA \cite{li2021transferable}, our domain augmentation method achieves performance improvement of $0.8\%$ on average. 


\textbf{Office-31} The results on Office-31 with ResNet-50 are presented in Table \ref{tab:Office-31}. Compared to the baseline methods, DADA achieves higher performance on tasks D $\rightarrow$ A, W $\rightarrow$ A, as well as the average performance, with the accuracy of $76.5\%$, $75.0\%$ and $89.6\%$ respectively.

\textbf{VisDA-2017} The results on VisDA-2017 dataset with ResNet-101 are summarized in Table \ref{tab:visda}. In VisDa-2017 dataset, the data from the source domain are relatively disparate from the data in the target domain. In some specific categories such as truck, skate board, and knife, DADA achieve an improvement of from $45.5, 68.9, 79.6$ to $52.8, 82.3, 84.1$ respective, compared to the second-best performing model. Although in some aspects, DADA performs inferior than other models, in general, we can observe form the results that DADA achieves high performance on this large dataset. 

\textbf{Digital datasets} The results on Digital dataset are summarized in Table \ref{tab:visda}. In average, our DADA method also achieves state-of-the-art performance on the Digital dataset. Digital dataset contains small datasets for digital image classification, where the domain gap between the source and target domain is not broad. Thus, the results indicates that DADA can not only be applied on large datasets such as Office-Home and VisDA-2017, but also works on small datasets which can be related to many simple use cases.


\subsection{Ablation Study \& Discusses}

In this ablation study, we aim to analyse the effectiveness of each individual component of DADA, including pseudo-domain augmentation, target labeling from pseudo domains, domain discrepancies losses, self-supervised losses and domain consistency losses. We conduct the ablation studies on the Office-31 dataset with ResNet-50 as our backbone. The corresponding results of ablation studies are shown in Table \ref{tab:ablation}.

\textbf{Effectiveness of pseudo domains.} The first row of Table \ref{tab:ablation} reflects the result of removing pseudo domain generation, by directly minimizing the source and target domain and the target domain, where we also add pseudo labels technique to enhance the domain adaptation. The second row shows the results of applying pseudo domains (with domain consistency loss) together with the pseudo labels that are generated from the pseudo domains. On average, purely applying multiple pseudo domains results in improvements of $3\%$, from $82.2$ to $85.5$.

\textbf{Effectiveness of pseudo domain separation.} In the third row, we include the pseudo domain separation loss according to Eq (\ref{eq:seperation}). With this component, the pseudo domains are pushed away from each other during the training phase. Therefore, each pseudo domain forms a specific distribution where we transfer knowledge from the overall pseudo domain representation space to the target domain. From the experiment results, we can observe that with the pseudo domain separation loss, the average accuracy on the target domain improves from $87.0$ to $88.6$. 

\textbf{Effectiveness of mapping (target to pseudo).} Furthermore, we analyze the effect of our proposed pseudo labeling method. Row five represents our proposed DADA method, and row four adopts the same components as DADA but only replaces our proposed pseudo labeling method by directly using the highest prediction on the target domain data as the pseudo labels. The results demonstrate the superiority of our proposed pseudo labeling method as this component itself achieves an average improvement from $88.6$ to $89.3$.



\textbf{Comparison with different number of pseudo domains.} To test the influence of different numbers of pseudo domains on the performance of DADA, we conduct experiments on Office-31 with the number of pseudo domains from 1 to 7. The corresponding results are shown in Table \ref{tab:pseudo_m}. The results show that the performance of DADA increases with the number of pseudo domains, however, the performance may be saturated when for a large number of pseudo domains, e.g. 4. 

\begin{figure}
\vspace{-0.5cm}
    \centering
    \includegraphics[width=4cm]{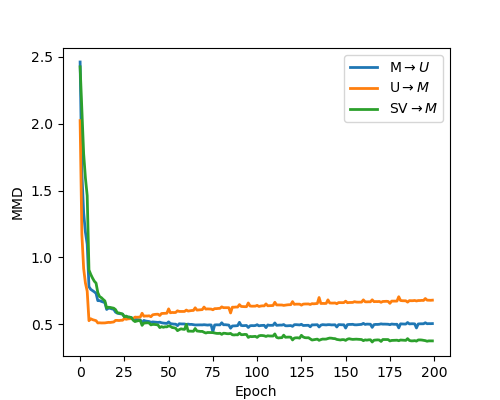}
    \caption{The average MMD between pseudo domains and the target domain during training on Digital datasets for M$\rightarrow$U, U$\rightarrow$M, and SV$\rightarrow$M. }
    \label{fig:mmdepoch}
\vspace{-0.5cm}
\end{figure}

\textbf{Discrepancy between pseudo domains and the target domain.} In addition, to test whether the pseudo domains can form similar distributions as the target domain during the training processes, we conduct extra experiments to test the average discrepancy between pseudo domains and the target domains at each training stage (i.e. epochs). We apply the MMD as the measure of the domain discrepancy. The results on Digital datasets are shown in Figure \ref{fig:mmdepoch}. The results show that the MMD between the pseudo domains and the target domains converges fast, and indicate that the adaptation from pseudo domains to the target domain can be achieved in a few epochs.

\textbf{Efficiency.} Careful readers may notice that the main limitation of the DADA is that due to multiple pseudo domains, the training time may be slowed down. However, with an appropriate number of pseudo domains, e.g. 3 or 4, we can balance the tradeoff between accuracy and efficiency. We further discuss the training time problem in the Appendix. 

\section{Conclusion}

In conclusion, in this work, we present the DADA method for unsupervised domain adaptation. DADA method generated multiple pseudo domains which have less discrepancy to the target domain compared to the source domain. To conduct domain adaptation, we minimize the discrepancy between the target domains and the pseudo domains. Meanwhile, we also design a novel pseudo labeling method assigning pseudo labels based on an average prediction of the pseudo domains. The extensive experiments were conducted and the results show that our proposed DADA method can achieve state-of-the-art performance in the domain adaptation. The ablation studies also demonstrate the effectiveness of each component of DADA. For future works, one of the main directions is to design better regularization methods that can further restrict the representation space of the pseudo domain for better adaptation. 




{\small
\bibliographystyle{ieee_fullname}
\bibliography{egbib}
}
\newpage
\clearpage



\appendix

\section{Visualization}
\label{sec:intro}

We visualize the pseudo domains through a reverse mapping network which maps the latent representations $Z$ to the data space, similar to the \textit{reverse mapping algorithm} described in TSA \cite{li2021transferable}. Taking the domain adaptation task SVHN $\rightarrow$ MNIST, our algorithm generates images in the pixel-level space corresponding to the pseudo domain features (i.e. SVHN $\rightarrow$ pseudo domain $\mathcal{D}_\gamma$). To utilize the reverse mapping algorithm in DADA, we firstly train Generative Adversarial Networks (GANs) with the data from the target domain (i.e. MINST), where the generator $G$ of the GANs learns to generate fake images based on the noise representation $z$, where $z \in \mathbb{R}^d$ denotes the random noise.

The goal of our reverse mapping algorithm is to find the desirable optimal noise $z^*$, such that 

    

\begin{equation}
\boldsymbol{z}^{*}=\arg \min _{\boldsymbol{z}}\left\|F(G(\boldsymbol{z}))-f_{g}(F\left(\boldsymbol{x}_{s}\right), \gamma)\right\|_{2}^{2},
\end{equation}

where $F$ denotes our backbone model (i.e. ResNet) that extracts features from the input images and $f_{g}$ denotes the pseudo domain generator with a pre-defined $\gamma$. Here, we randomly select a suitable $\gamma$ such that for the trained DADA model, the domain gap between $\mathcal{D}_\gamma$ and the target domain is relatively small. $x_s$ denotes the source data. Then, we visualize the pseudo domain representation of $x_s$ by generating the data via $G(z^*)$. The corresponding results are shown in Figure 2. In general, the visualization shows that the pseudo domain can reconstruct meaningful data through its representations, which also reflects the suitability of the pseudo domain representations.


\begin{figure}[h]
    \centering
    \includegraphics[width=8cm]{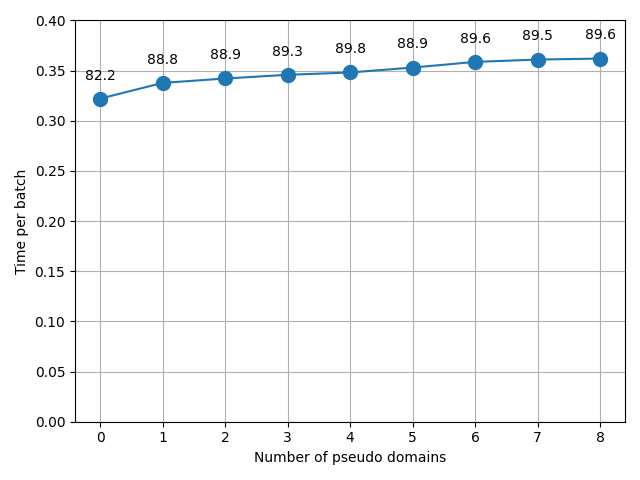}
    \caption{Experiment on the efficiency analysis. We conduct the experiment on the Office-31 dataset with the batchsize of 128. The y-axis represents the time consumption per batch in seconds, and the x-axis represents the number of pseudo domains ranging from zero to eight. The average accuracy of each setting is also shown right above each result point.}    \label{fig:my_label}
\end{figure}

\section{Efficiency Analysis} Careful readers may notice that the efficiency of applying multiple pseudo domains might be a major limitation of the DADA, as the training time should be slowed down with the increase of pseudo domains. To analyze the consumption of applying multiple pseudo domains, we conduct experiments on Office-31 dataset with the number of pseudo domains ranging from zero to eight. Zero pseudo domain represents the case when the DADA is not used, but we use the MMD to minimize the source and target domain gap instead.

The corresponding results are shown in Figure 1. The y-axis shows the time consumption per batch in seconds, where we set the batchsize of 128. The x-axis represents the number of pseudo domains. For each setting with different number of pseudo domains, we annotate the corresponding averaged accuracy for the DADA on Office-31. In general, we can infer that the pseudo domain indeed takes up a relatively small portion of the whole training time in a batch. Even for the eight pseudo domains setting, the portion of the time to generate and train pseudo domains only takes up $12.4\%$ of the entire batch time. Furthermore, applying four pseudo domains achieves the highest accuracy in this domain adaptation task, where DADA only takes additional $8.1\%$ computation time compared to the MMD model.

\begin{figure*}[t!]
    \centering
    \includegraphics[width=15cm]{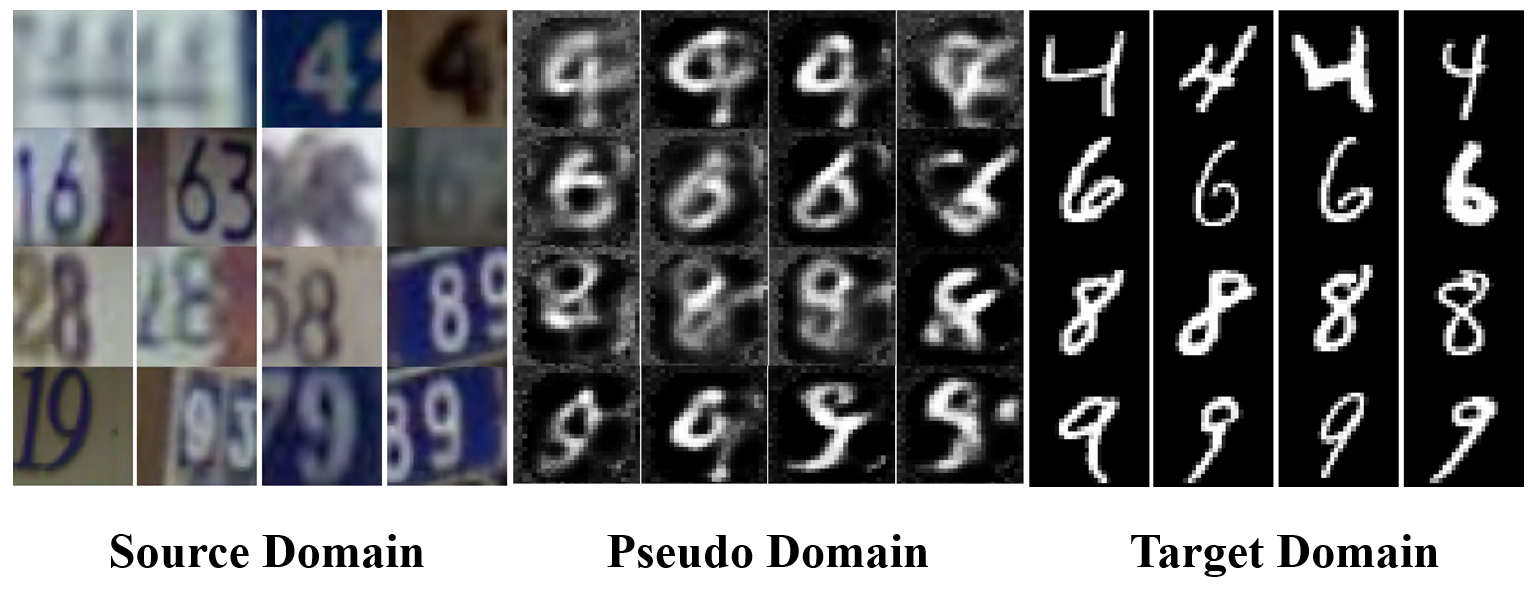}
    \caption{Visualization of source domains, representations from the pseudo domains and the target domain. Each row of the data corresponds to the same label.}
    \label{fig:my_label}
\end{figure*}

\end{document}